\documentclass{article}

\usepackage{arxiv}

\usepackage{graphicx} 
\usepackage{lineno}
\usepackage{url}
\usepackage[breaklinks]{hyperref}
\usepackage{booktabs}
\usepackage{tabularx,longtable,multirow,subfigure,caption}
\usepackage{amsmath}
\usepackage{textpos}
\usepackage{combelow}
\usepackage{xcolor}

\modulolinenumbers[5]

\usepackage{amssymb}
\usepackage{amsmath}
\usepackage{dsfont}
\usepackage[ruled,vlined]{algorithm2e}

\newcommand{\R}[0]{\mathds{R}} 
\renewcommand{\vec}[1]{{\boldsymbol{{#1}}}} 
\newcommand{\mat}[1]{{\boldsymbol{{#1}}}} 

\title{A novel approach for predicting epidemiological forecasting parameters based on real-time signals and Data Assimilation}

\author{Romain Molinas \\
	Department of Computing, \\ 
    Imperial College London, UK \\
	\And
	César Quilodrán Casas \\
	Data Science Institute, \\
    Imperial College London, UK \\
    \And
	Rossella Arcucci \\
	Data Science Institute, \\
    Department of Earth \\ Science \& Engineering, \\
    Imperial College London, UK \\
    \And
	Ovidiu \cb Serban \\
	Data Science Institute, \\
    Department of Computing, \\ 
    Imperial College London, UK
}

\date{}

\begin{document}
\maketitle

\begin{abstract}
This paper proposes a novel approach to predict epidemiological parameters by integrating new real-time signals from various sources of information, such as novel social media-based population density maps and Air Quality data. 
We implement an ensemble of Convolutional Neural Networks (CNN) models using various data sources and fusion methodology to build robust predictions and simulate several dynamic parameters that could improve the decision-making process for policymakers. Additionally, we used data assimilation to estimate the state of our system from fused CNN predictions. The combination of meteorological signals and social media-based population density maps improved the performance and flexibility of our prediction of the COVID-19 outbreak in London.
While the proposed approach outperforms standard models, such as compartmental models traditionally used in disease forecasting (SEIR), generating robust and consistent predictions allows us to increase the stability of our model while increasing its accuracy. 

\medskip\noindent\textbf{Highlights:}

\begin{itemize}
    \setlength\itemsep{0em}
    \item A novel model to predict epidemiological parameters of COVID-19, such as the number of infections and deaths.
    \item The proposed models are based on a flexible fused architecture which allows easy swap of individually trained models.
    \item Integrating social-media data as a real-time signal for population behaviour in a densely populated area.
    \item Using meteorological and air quality metrics to characterise epidemiological spread in London.
\end{itemize}
\end{abstract}


\section{Introduction}

In epidemiological forecasting, traditional models and measurements take days to collect, aggregate, and integrate into existing statistical models. These predictions are extremely important in understanding the geographical spread of disease and help authorities in the decision-making process \cite{lutz2019applying}. We propose a novel approach to predict some of the epidemiological parameters, such as the number of new infections and deaths, by integrating new real-time signals that are traditionally not considered by experts in this field. Moreover, to compensate for the uncertainty, we apply Machine Learning (ML) and Data Assimilation (DA) techniques by integrating observations into our forecasting model, which tend to increase the model stability and decrease the error rate. 

Constraints in standardised case definitions and timely and noisy data sources can affect the accuracy of predictive models. Due to the lack of granular data, resource-limited environments pose specific challenges for accurate disease prediction \cite{DESAI2019}. Therefore, epidemiological forecasting using predictive modelling is an effective resource for anticipating outbreaks and tailoring responses.

Over the past two decades, various approaches to epidemiological modelling have been explored to provide sufficient resources for understanding and analysing outbreaks \cite{WALTERS20181}. Spatial information has successfully improved the understanding of epidemic dynamics by applying various spatial analytics, such as geographical information, residential mobility, sensed environmental data, and spatio-temporal clustering \cite{KIRBY20171, BROOKER20071}. Moreover, compartmental models have proved very effective in emulating epidemic outbreaks through a set of differential equations. Optimal solutions of these deterministic equations can reasonably infer the inner mechanisms of an epidemic outbreak, such as the spread of the disease \cite{VANDENDRIESSCHE200229}. 

The introduction of government action parameters in the transmission rate, such as social and political measures, improved the modelling of the 1918 influenza pandemic in the UK \cite{flu_uk} and the recent COVID-19 pandemic outbreak in Wuhan, China \cite{LIN2020211}. The prediction of the transmission rate has also shown potential for forecasting based on parameter estimation \cite{SMIRNOVA2019, nadler2020epidemiological, wang2020bayesian}. 
Recent bio-surveillance systems were able to detect earlier disease outbreaks on real-time noisy social media data \cite{SERBAN20191166, GHIASSI20136266}.

In this paper, we present our implementation of a combination of CNN models from various data sources to build robust predictions. Our fused CNN architectures track spatial and temporal relationships across the real-time data stream and thus emulate the dynamic parameters. Then, our approach's performance and flexibility are assessed on the COVID-19 outbreak using temporal and spatial public information sources from various official institutions, social media and air quality data. Finally, we expose the benefit of DA to robustly estimate the system's state with the fused CNN and its consistent prediction.

\section{Related Work}

The accuracy of prediction models can be affected by constraints in conventional case definitions and timely and noisy data sources. Resource-constrained areas present unique obstacles for reliable emerging epidemic prediction caused by a lack of actionable insights \cite{DESAI2019}. The inherent uncertainty, not just about the contagious virus itself but also about the interconnected human, societal, and political aspects that co-evolve and keep the future of outbreaks open-ended, makes projecting future events in the pandemic difficult \cite{LUO2021120602}. 

Over the past two decades, various approaches to epidemiological modelling have been explored to provide sufficient resources for understanding and analysing outbreaks\cite{WALTERS20181}.

Compartmental models, a population-wide established approach, have been proven to be very effective in emulating epidemic outbreaks by modelling individuals as a finite number of discrete states. The states reflect the progression of an infection with a time dependency: a person may become infected, recover or deceased. 

The dynamics of the states are governed by a parametric nonlinear ordinary differential equation (ODE). Optimal solutions of these deterministic equations can reasonably infer the inner mechanisms of epidemic outbreaks and provide essential quantities of infectious disease outbreaks \cite{refId0}. 

The compartmental model offers high dimensional systems able to precisely discretise populations into groups of individuals with similar disease infectious dynamics. However, the spread of infectious diseases can be influenced by inherent virus properties, human characteristics (e.g. ages, sex, genome, genetic diseases), environment (e.g. hygiene, culture, household) and mobility (e.g. local mobility on daily journeys, national and international mobility)\cite{LIN2020211}, meteorological factors \cite{RENDANA20211320}  and probably other unknown factors might also affect the dynamics of the spread. Consequently, integrating those factors increases the complexity of the ODE for a profit of information gain, difficult to evaluate and optimise by estimating appropriate parameters for the equation especially based on a real-time stream of information. 

Recently, a first step toward integrating these factors into disease model parameters has been made. In particular, spatial information has successfully improved the understanding of epidemic dynamics by applying various spatial analytics, such as geographical information, residential mobility, sensed environmental data, and spatio-temporal clustering \cite{KIRBY20171, SLATER2021100540}. The prediction of the transmission rate has also shown potential for forecasting based on parameter estimation \cite{SMIRNOVA2019}. Activation of this knowledge opens up an interesting field of possibility to create robust and efficient epidemic prediction models\cite{ARUNKUMAR2021110861}.

Simultaneously, social media data has also gained attention over the past few years for its enormous wealth of information and knowledge encapsulated. Social Media and messaging apps are the most common communication methods. Advances have shown that it is possible to extract valuable knowledge from the colossal amount of information and the saturation of content and to systematize complex decision-making processes in the service of the marketing and commercial strategies of companies\cite{YANG2022102751}. 
The potential of social media-based analysis tools has begun to be explored to help decision-making during crises, especially health crisis. 
In the same direction, Facebook proposed Disaster Maps for crisis analysis and response. They are built from comparing count-based of a certain event during crisis and expectations from pre-crisis periods based on a two-weeks aggregation of anonymous data from Facebook users \cite{MAAS2019FB}.
Recent bio-surveillance systems could also detect earlier disease outbreaks using DA on real-time noisy social media data. A software framework based on modular model processing real-time data and using deep learning for the classification of health-related tweets have had encouraging results for forecasting influenza outbreaks in some regions of the USA\cite{SERBAN20191166}. On the same track, in the research work of diagnosing clinical infection at early stages \cite{BALDOMINOS2020102213}  from a local small amount of patient data, the use of social data and weather data readily available increased the performance of the prediction by 9\%.

Motivated by the latest outbreak of COVID-19, the analysis of the public's reaction and their perception of information during emerging infectious diseases based on emotion intensity, message volume \cite{QIU2020103217} and content have improved the ability to capture the acceptance and adoption of sanitary measures and related compartment \cite{CHAWLA2021102720}. 
 Research efforts have been consequent in building a robust and accurate epidemic model. In this regard, a  novel data-driven epidemic model has been developed, which focuses on marked temporal point processes precisely engineered to allow fine-grained spatio-temporal estimates of the spread of the disease within a population \cite{lorch2021quantifying}. They show unprecedented spatio-temporal resolution to quantify the effects of tracing, testing, and lockdown.
 In this context, the cyclical nature of the phases of epidemics has motivated research to capture the underlying dynamics. Several models using autoregression approaches have emerged. They relied on the prediction of COVID-19 dynamics: the number of  confirmed cases, confirmed recovered and deaths using ARIMA with added weather data \cite{RENDANA20211320}, Seasonal model \cite{ARUNKUMAR2021107161}. The heterogeneous autoregressive model preferred to capture long-memory features from data better \cite{HWANG2021104631}. Deep learning model, RNN and LSTM more precis have also been tested to predict dynamics of an epidemic using a trade-off between long and short memory from the multiple sources of data \cite{ARUNKUMAR2021110861, QUILODRANCASAS202211}.
 
However, until the new trend and a stationary situation are picked up, ARIMA and another autoregressive model with statistical curve fitting approaches should generally show very poor performance in periods where there is abrupt and non-stationary growth in the number of active COVID-19 cases that cannot be attributed to "seasonal" components. Also, individual model predictions have been demonstrated to be highly sensitive to parameter assumptions according to the research work performed through multiple COVID-19 model predictions applied in ten countries \cite{DREWS2022150639}. An ensemble model approach demonstrated higher robustness and flexibility by reducing uncertainties from various data sources capturing constantly changing dynamics of COVID-19 transmissions at local levels.

From the literature review, these advances show the opportunities to create more robust and efficient data-driven models for epidemic outbreak forecasting. But most of these models are still highly sensitive to parameter estimation, impacting the performance of epidemic dynamics forecasting when lacking knowledge about health situations when outbreaks occur or when changes in behaviours are induced by sanity measures, for instance (lockdown, vaccination, ...). 

\section{Research Objectives}
Our approach focuses on creating a real-time data-driven model to emulate the dynamics of epidemic outbreaks. The characteristics of our model include flexibility and robustness to process a wide range of sources of potentially noisy real-time signals.
The objective is to build a model capable of selecting and processing the most useful information to simulate the underlying dynamics of an outbreak system by automatically activating and adjusting parameters to maximise the information gain.
The information is processed to evaluate the population environment in a dynamic time and space manner to offer knowledgeable features and insights for our model to forecast epidemiological parameters precisely. 

Our main objectives are listed below :
\begin{itemize}
    \setlength\itemsep{0em}
    \item A novel model to predict epidemiological parameters of COVID-19, such as the number of infections and deaths.
    \item The proposed models are based on a flexible fused architecture which allows easy swap of individually trained models.
    \item Integrating social-media data as a real-time signal for population behaviour in a densely populated area.
    \item Using meteorological and air quality metrics to characterise epidemiological spread in London.
\end{itemize}

Our research focuses on combining observational data with output from a numerical model to obtain an ideal estimate of the evolving state of our system to bring robustness and resistance to real-time signals.

\section{Proposed Method}
\subsection{Combining Real-Time Signals Forecast Models}\label{sec:fusedcnn}

\begin{figure}[htb]
    \centering
    \includegraphics[width=0.30\linewidth]{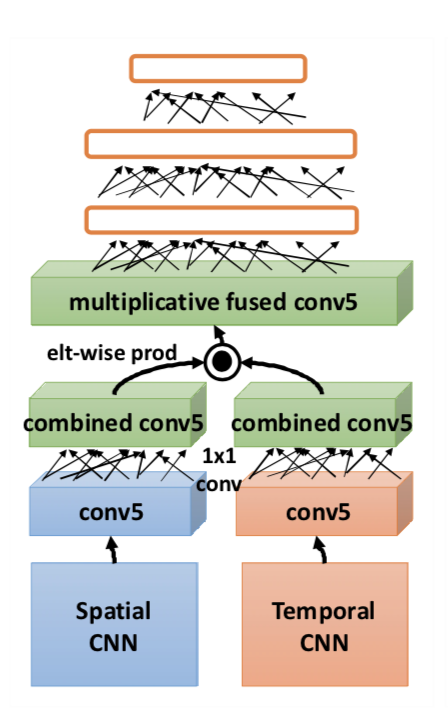}
    \caption{CNNs fusion architecture proposed by \cite{PARK2016}}
    \label{fig:network-fusion-park-2016}
\end{figure}

As inspired by the practicality of the CNN fusion model proposed by \cite{PARK2016}, our approach uses a multiplicative fusion method by combining several CNNs trained on different sources, resulting in robust prediction by amplifying or suppressing a feature activation depending on their agreement, as shown in Figure \ref{fig:network-fusion-park-2016}. 

The proposed approach is intended to identify each network's important features and give a higher prediction score only when several networks agree with each other.
For instance, if both networks generate high activations at the same location, activations of feature maps from the convolutional layer will be intensified otherwise, they will be discarded.

To formulate the fusion method, we can consider two CNNs: temporal and spatial networks. From each network, let's consider the matrices $A \in \R^{d \times M}$ and $B \in \R^{d \times N}$ to be extracted features of the last convolutional layer. $M$ and $N$ are the numbers of feature maps based on CNNs architecture, and $d$ is the size of feature maps.
The output of the fusion method is :

\begin{align}
    \vec{c_k}=\left (\sum_{i=1}^{M}\alpha_{ki} \vec{a_{i}} + \gamma_k \right) \odot \left (\sum_{j=1}^{N}\beta_{ki} \vec{b_{i}} + \delta_k \right)
\end{align}

$\vec{a_{i}}$ and $\vec{b_{i}}$ are the $i$th column of each matrix, which corresponds to one feature map. The operator $\odot$ is the element-wise product, and $\gamma$ and $\beta$ are bias terms. $\alpha$ and $\beta$ are weights for each feature map and learnable parameters. They are preponderant in selecting good features in each network, giving higher weights to useful features for prediction. This layer is trained with standard back-propagation and stochastic gradient descent.

In our work, we proposed an extension of the above fusion method by adding extra hyperparameters that can be fine-tuned during training and feeding the results into another Neural Network to predict epidemic outcomes. Our proposed fusion ingredients are two CNNs: temporal CNNs and spatial CNNs, respectively trained on the temporal stream of information and a spatial stream of information.

\begin{figure}[htb]
    \centering
    \includegraphics[width=\linewidth]{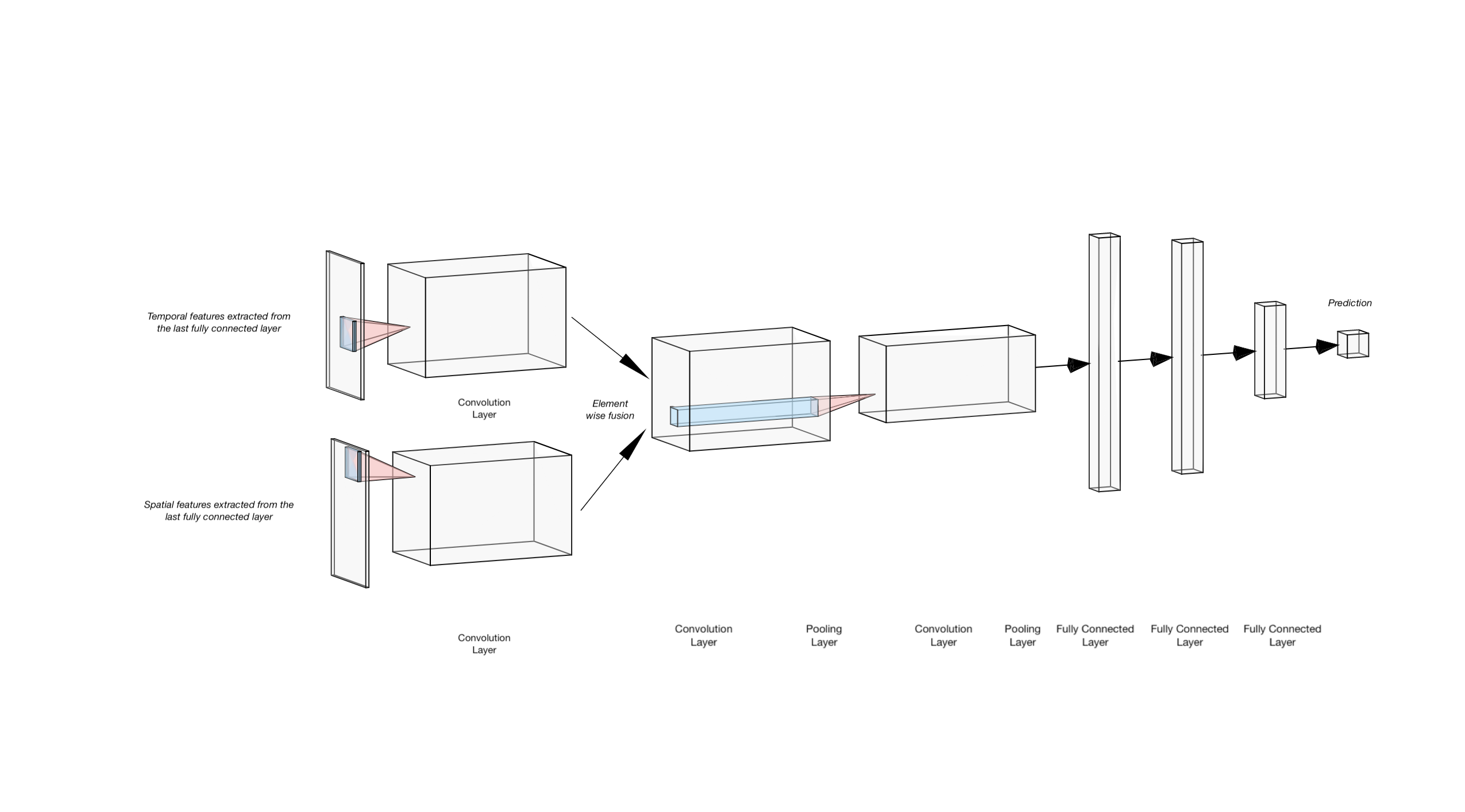}
    \caption{The combination of deep learning models is performed by element-wise multiplication between the feature maps extracted from specific CNN models} 
    \label{fig:fused-architecture}
\end{figure}

All the models are pre-trained on their respective datasets.  Then, weighted sums of feature maps from the latest convolutional layer of both networks are computed. These networks are then joined by element-wise multiplications into a convolutional layer with fully connected layers to perform the regression. The proposed layers are trained with standard back-propagation and optimised using the Stochastic Gradient Descent (SGD). This method provides different features extracted for every tracked dataset. The combination of the CNNs outputs allows cross-communication between networks and consequently enhanced knowledge from other streams. The full architecture is shown in Figure \ref{fig:fused-architecture}.

This approach has been preferred compared to a full CNN model because the reduced size of the sub-models significantly decreases computational complexity. If required, the re-training is also faster, which is highly desired for real-time forecasting. Additionally, multiple sub-models provide a simpler understanding of each feature extracted and an easy explanation of the inferred influence of mobility and population in the parameters of the epidemic.

\subsection{Improving the forecasting using Data Assimilation}
\label{sec:data-assimilation-theory}

Our data-driven approach is composed of several time series of multi-dimensional observations $\textbf{y}_k \in \R^p $ of an unknown process $\textbf{x}_k \in \R^m $: 

\begin{align}
    \textbf{y}_k = H_k( \textbf{x}_{k} )  + \epsilon_k^y
\end{align}

Observations are indexed from $0 \leq k \leq K $ where k corresponds to the time $t_k$. $H_k$ represents the observation operator: $H_k :\R^m \rightarrow \R^p$. We assume that the observation error is known and represented by $\epsilon_k^y$  where the assumption of following a normal distribution of mean and covariance matrix $\mat{R}_k$ hold.

We suppose that the process $\textbf{x}$ obeys partial differential equations and the prediction of the following state at time $t_{k+1}$ given the state $t_k$ is defined :

\begin{align}
    \textbf{x}_{k+1} = G_k( \textbf{x}_{k} )  + \epsilon_k^x
    \label{eq:surogate-model}
\end{align}

where $\epsilon_k^m$ is the error of the surrogate model of the PDE mentioned in equation (\ref{eq:surogate-model}) following a Gaussian distribution with zero mean and covariance matrix $\mat{P}_k$. 

The random vectors $\epsilon_k^y$and $\epsilon_k^m$ represent the modelling and observation errors, respectively. They are assumed to be independent, white-noise processes with Gaussian/normal distribution :

\begin{align}
    \epsilon_k^y \sim \mathcal{N}(0,\,\mat{R}_k) \\
    \epsilon_k^x \sim \mathcal{N}(0,\,\mat{P}_k)
\end{align}

The representation of the surrogate model uses the deep learning step. It is composed of concatenated CNNs sub-models from the spatio-temporal stream of information. We represent our model with a parametric function $G_{\textbf{W}}(\textbf{x})$ :

\begin{align}
    G_{\textbf{W}}(\textbf{x}) = \textbf{x}_k + f_{CNNs}(\textbf{x}_k , \textbf{W})
    \label{eq:deep-learning-model}
\end{align}

The function $f_{CNNs}$ is a deep neural network pre-trained with its corresponding weights $\textbf{W}$.

\subsection{Stochastic Ensemble Kalman Filter}

\textbf{Analysis Step}: The goal of the stochastic EnKF is to perform for each member of the ensemble an analysis of the form \cite{book} :

\begin{align}
    \textbf{x}_{i}^{a} =  \textbf{x}_{i}^{f} + \mat{{K}_{i}}\left[\textbf{y} - H( \textbf{x}_{i}^{f})  \right]
\end{align}

where $i=1, \dots, k$ represents the member index in the ensemble and $x_{i}^{f}$ the forecast state vector $i$, defining the prior at the given analysis time.

$\mat{K}$ is identified with the Kalman gain: 
\begin{align}
    \mat{{K}_{i}} = \mat{P^f}\mat{H}^T( \mat{H}\mat{P^f}\mat{H}^T + \mat{R})^{-1}
    \label{eq:kalman-gain}
\end{align}

This quantity is estimated from the ensemble statistics. The forecast error covariance matrix is :

\begin{align*}
    \mat{P^f} = \frac{1}{m-1}\sum^{m}_{i=1}(\textbf{x}_{i}^{f} - \bar{\textbf{x}}^{f})(\textbf{x}_{i}^{f} - \bar{\textbf{x}}^{f})^T \\
   \bar{\textbf{x}}^{f} = \frac{1}{m}\sum^{m}_{i=1}\textbf{x}_{i}^{f}
\end{align*}

Using normalised perturbations from \cite{book}, the gain can be computed using only the anomaly matrices :
\begin{align}
 \mat{{K}_{i}}  = \mat{X_f}\mat{Y_f}^T(\mat{Y_f}\mat{Y_f}^T)^{-1} 
    \label{eq:anomaly-matrices}
\end{align}

where: 

\begin{align*}
       [\mat{X_f}]_i=\frac{\textbf{x}_{i}^{f} -\bar{\textbf{x}}^{f} }{\sqrt{m-1}} \\
     [\mat{Y_f}]_i=\frac{\mat{H}\textbf{x}_{i}^{f} - \textbf{u}_i -\mat{H}\bar{\textbf{x}}^{f} + \bar{\textbf{u}} }{\sqrt{m-1}}
\end{align*}

The perturbation $\textbf{u}_i$ (from the Gaussian distribution $\sim \mathcal{N}(0,\,\mat{R})$, where $\mat{R}$ is the covariance matrix) is added to the observation vector for each member of the ensemble. This provides a solution to a divergence of the EnKF generated from an underestimation of the error covariances \cite{book}. 

In equation \eqref{eq:anomaly-matrices} the term $\mat{X_f}\mat{Y_f}^T$ is a sample estimate for  $\mat{P^f}\mat{H}^T$, taken from equation \eqref{eq:kalman-gain}. Similarly, $\mat{Y_f}\mat{Y_f}^T$ is a sample estimate for $\mat{H}\mat{P^f}\mat{H}^T + \mat{R}$. In this shape, it is interesting that the updated disturbances are linear combinations of the forecast perturbations. New perturbations are being found within the subspace ensemble of the original perturbations.

\textbf{Forecasting Step}: In the forecasting step, the updated ensemble obtained during the analysis step is propagated by the Deep Learning model (equation \ref{eq:deep-learning-model}) over a time step and for all the particles of the ensemble $i=1, \dots,m$, where $G_k$ is defined by equation \ref{eq:deep-learning-model}:

\begin{align*}
    x_{i,k+1}^f= G_k( \textbf{x}_{i,k}^a )  
\end{align*}

In our work, the forecast error covariances are estimated from the forecast perturbations. Regarding the forecast, two different computations have been explored: the forecast ensemble's mean and the forecast ensemble's median. With the significant exception of the Kalman gain measurement, all operations of the ensemble members are independent, which means that all the training can be done in parallel.

The use of EnKF may be a bit excessive for the amount of data available in our case. However, we built this approach as a flexible data-driven methodology capable of self-adaptation-based availability and granularity of data. More importantly, EnKF filtering is able to incorporate the uncertainty of input parameters used in model forecasting the spread of diseases \cite{DESAI2019}. Considering the noise in our ground truth data and the uncertainty around the collection, we strongly believe the DA improves the stability of our results. 

The sequential Bayesian filter implements an ensemble of state vectors to represent the distribution of the system states vector $\vec{x}$. Considering the choice of ground truth data, the states that can be potentially assimilated are the daily deaths and the daily lab-confirmed cases. This means that the other features should be concatenated with the assimilated states to forecast the next state, as DA requires the previous state of the forecasting as input.

The implementation of the EnKF is inspired by the Stochastic EnKF algorithm proposed by \cite{book}. Our adaptation of this algorithm is shown by Algorithm \ref{algo1}. The observation model operator is the identity matrix $H \in \R^{dim(x)\times dim(x)} $, size of the length of the system states vector $x$. The forward model operator is the fused CNN represented with the parametric function $G_{\textbf{W}}$. All the parameters from the combined CNNs and their respective hyperparameters are included inside the $\textbf{W}$ term. The selection operator $S$ is providing an estimation of the ensemble forecast. Simple functions are considered to be experimented such as the mean computation or the median.

\begin{algorithm}[h!tb]
\SetAlgoLined \DontPrintSemicolon
\SetKwInOut{Input}{Require}\SetKwInOut{Output}{Result}
\Input{For $k=0, ... K $: the observation error covariance matrices $\textbf{R}_k$, the observation models $H_k$, the forward models $G_{\textbf{W}}$}
\Output{$\Big\{\textbf{x}^{f}_{i,0}\Big\}_{K}$, the next states at step $K$}
\;
\tcc{Initialise the ensemble}
$\Big\{\textbf{x}^{f}_{i,0}\Big\}_{1,...,m}$ = 0 \;
\BlankLine
\For{k=0,...,K}{
  \tcc{Generate statistically consistent observation set :}
  \For{i=1,...,m}{
  $\textbf{y}_{i,k}=\textbf{y}_k + \textbf{u}_i$ , with $\textbf{u}_i \sim \mathcal{N}(0,R_k)$
  }
  
    \tcc{Compute the ensemble means:}
    $\overline{\textbf{x}_k}^f=\frac{1}{m}\sum_{i=1}^{m} \textbf{x}_{i,k}^f$\;
    $\overline{\textbf{u}}=\frac{1}{m}\sum_{i=1}^{m} \textbf{u}_{i}$\;
    $\overline{\textbf{y}_k}^f=\frac{1}{m}\sum_{i=1}^{m} H_k( \textbf{x}_{i,k}^f)$\;
    \;
    \tcc{Compute the normalised anomalies:}
    $[\textbf{X}_f ] _{i,k}=\frac{\textbf{x}_{i,k}^f - \overline{\textbf{x}_k}^f}{\sqrt{m-1}}$ \;
    $[\textbf{Y}_f ] _{i,k}=\frac{H_k( \textbf{x}_{i,k}^f)- \textbf{u}_i -\overline{\textbf{y}_i}^f + \overline{\textbf{u}} }{\sqrt{m-1}}$ \;
    \;
    \tcc{Compute the gain:}
    $\textbf{K}_k = \textbf{X}_k^f(\textbf{Y}_k^f)^T\Big\{\textbf{Y}_k^f( \textbf{Y}_k^f)^T\Big\}^{-1}$\;
    \;
    \tcc{Update the ensemble:}
  $\textbf{x}_{k}^a = \textbf{x}_{k}^f +\textbf{K}_k( \textbf{y}_{k} - H_k( \textbf{x}_{k}^f))$ \;
    \;
    \tcc{Add the assimilated state values:}
    $\textbf{x''}_{k}^a= \textbf{x}_{k}^a + \textbf{x'}_{k}^a$\;
    \;
    \tcc{Compute the ensemble forecast:}
    $\textbf{x}_{k+1}^f = G_{\textbf{W}, k+1}(\textbf{x''}_{k}^a) $\;
    \;
     \tcc{Compute an estimation of the ensemble states using the selection operator $S$:}
    $\textbf{x}_{k}^f = S_k(\textbf{x}_{k}^f)$}\;
    \caption{Our adaptation of the Stochastic EnKF presented in \cite{book}}
    \label{algo1}
\end{algorithm}

\begin{figure}[htb]
    \centering
    \includegraphics[width=0.5\linewidth]{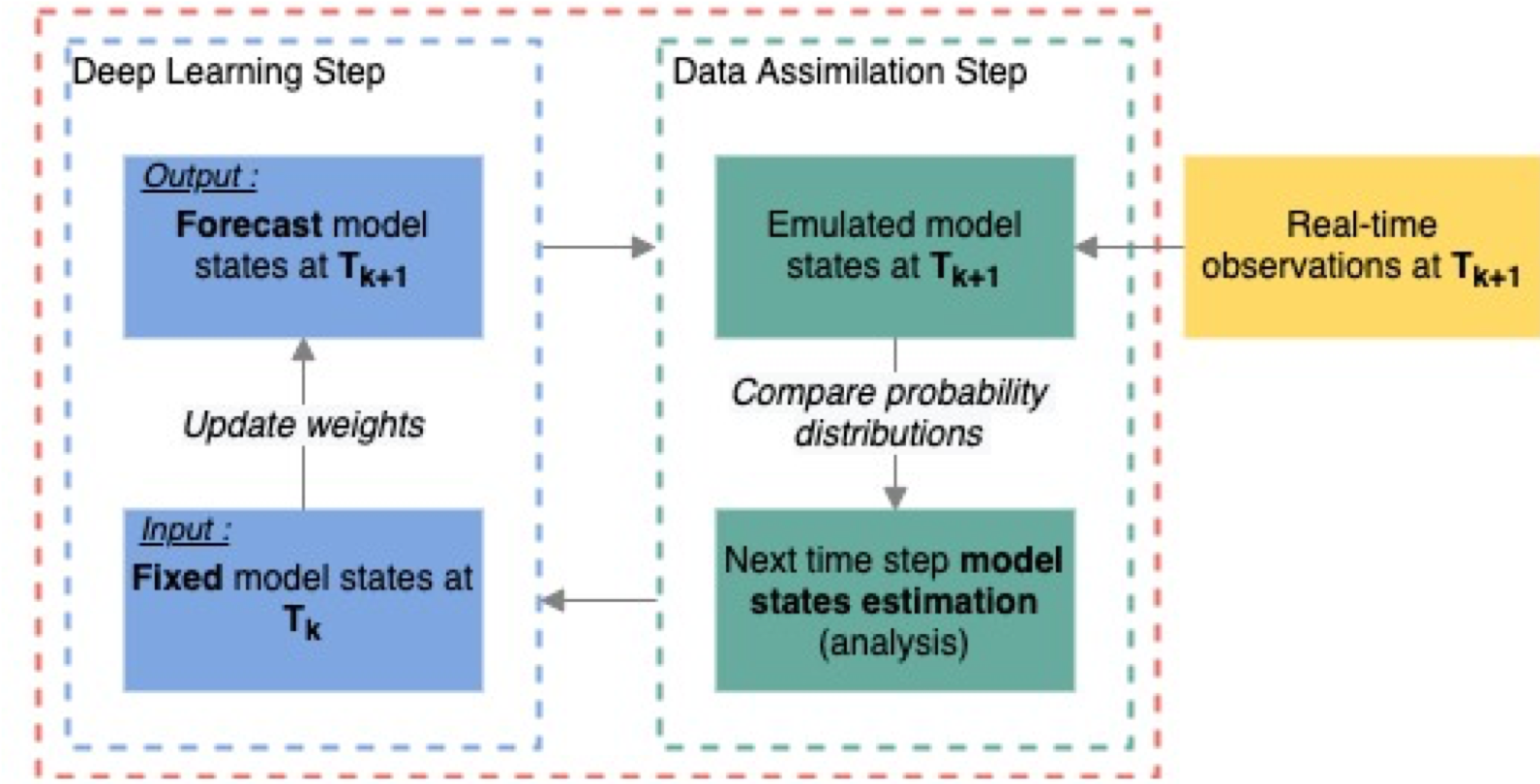}
    \caption{At each step, the Deep Learning  process provides DA with a surrogate forward model, and reciprocally, DA provides a time series of assimilated states to train the CNNs models}
    \label{fig:combined-cnn-da}
\end{figure} 

With the significant exception of the Kalman gain measure, all operations of the ensemble members in Algorithm \ref{algo1} are independent, which means that the algorithm can be easily executed in parallel or distributed across a cloud environment.

The DA process estimates the state of the system while the combination of Deep Learning models emulates the parameters of the dynamic model \cite{brajard2020combining}. The forecast probability is compared to the observations ensemble and produces the analysis. The output analysis is fed into CNNs to update the surrogate model. This iterative cycle is shown in Figure \ref{fig:combined-cnn-da}.

\section{Empirical Analysis }
\label{sec:experiements}

\subsection{Datasets}\label{sec:datasets}

Novel data streams, such as epidemic case incidence data provided by digital disease detection tools, demographic data estimates aided by geospatial mapping tools, and advances in mathematical modelling, can support efforts to control emerging outbreaks. It can also provide useful input to public health authorities complementing existing data sources and monitoring structures. Predictive models may exploit these innovative data sources to include timely case count forecasts and potential geographic spread of an evolving outbreak in real-time point out \cite{DESAI2019}.

 Although conventional disease surveillance remains the bedrock of the epidemic evaluation and regular data collection, informal disease surveillance systems enable faster dissemination and identification of case occurrence data. More importantly, emerging data sources will provide valuable insight for epidemic models in areas that do not have public health networks in place due to violent conflict or inadequate infrastructure. It also has benefits in data collection at the global level, since data is obtained using a standardised format \cite{DESAI2019}.

Population maps are provided by Facebook\footnote{More details at: \url{https://dataforgood.fb.com/docs/high-resolution-population-density-maps-demographic-estimates-documentation/}} have been exploited to infer spatial dynamics of the pathogen \textit{Facebook Population Map (Tile Level)}. These are location density maps or heat maps, which show where people are located before, during and after a disaster and where populations have increased or decreased. We can compare this information to historical records, such as population estimates based on satellite images. Comparing these data sets can help response organisations understand areas impacted by a natural disaster.

Recent medical studies show a strong correlation between the air quality metrics and risk of death \cite{pozzer2020regional}, therefore in our experiments, we use weather and air quality levels in the London area. 
 To get as close as possible to real-time processing, we decided to focus on data with the highest frequency. Simultaneously, we wanted to combine both knowledges from the temporal information and from the spatial information to infer as much as possible the parameter of the disease dynamics. Thus, we have selected \textit{London Air Pollution} data to cross-correlate these potential environmental drivers of epidemics and the spread of certain pathogens.The \textit{London Air Pollution} database\footnote{Available online at \url{http://www.londonair.org.uk/london/asp/datadownload.asp}) or through the Openair API.} provides atmospheric data measurements from various local authorities across the London region. This stream of information is subject to noise from the method of acquisition and its pre-processing by the \textit{London Air Pollution} teams. To cope with missing data, we use nearest neighbour interpolation. The nearest neighbour algorithm selects the value of the nearest point without considering the values of neighbouring points at all, resulting in piece-wise constant interpolation. 
 
The feature selection has been made keeping sight of our project objective of developing an approach capable of processing real-time signals from various public sources able to detect patterns in disease dynamics.
Two datasets are considered to run our experiments: A temporal stream and a spatial stream. The dataset for the temporal stream is composed of meteorological information :
 
\begin{itemize}
    \setlength\itemsep{0em}
    \item Barometric Pressure: measurement of air pressure in the atmosphere (mbar).
    \item Solar Radiation: the power per unit area received from the Sun in the form of electromagnetic radiation (W/m2).
    \item Temperature: measurement of how fast the atoms and molecules of a substance are moving (Celsius)
    \item Wind Speed: the speed of the weather-related air movement from one place to the next (m/s)
    \item Relative Humidity: the amount of water vapour actually in the air (percentage)
    \item PM10 \& PM2.5 Particulates: the number of particulate matters 10 \& 2.5 micrometres or less in diameter for a cubic meter of air, PM2.5 is generally described as fine particles (ug/m3)
    \item Carbon Monoxide: the amount of carbon monoxide particles for a cubic meter of air (mg/m3)
    \item Nitric Oxide: the number of nitric oxide particles for a cubic meter of air (ug/m3)
    \item Nitrogen Dioxide: the amount of nitrogen dioxide particles for a cubic meter of air (ug/m3)
    \item Oxides of Nitrogen: the number of oxides of nitrogen particles for a cubic meter of air (ug/m3)
    \item Ozone: the number of ozone particles for a cubic meter of air (ug/m3)
    \item Sulphur Dioxide: the number of sulfur dioxide particles for a cubic meter of air (ug/m3)
\end{itemize} 

During the research, we discovered that information from multiple public institutions is available. The \textit{Department of Health and Social Care} (DHSC)  and \textit{Public Health England} (PHE) have developed a dashboard since the COVID-19 outbreak in March\footnote{available at \url{https://coronavirus.data.gov.uk/}}, where daily case count and daily deaths are reported. This is our ground truth data :
\begin{itemize}
    \setlength\itemsep{0em}
    \item Daily and cumulative lab-confirmed cases:  the number of individuals with a lab-confirmed positive COVID-19 antigen test  on  or  before  the  sampling  date  or  reporting  date, published by \textit{Public Health England} (PHE)
    \item Daily Death: number of deceased hospitalised in England or had either tested positive for COVID-19 or where COVID-19 was mentioned on the death certificate.  All counts are recorded against the date of death rather than the announcement date.
\end{itemize} 

    \begin{figure}[htb]
    \centering
    \includegraphics[width=0.65\linewidth]{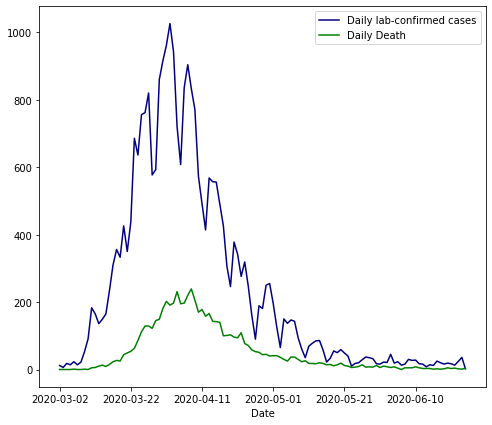}
    \caption{Ground truth data: Daily Numbers of lab-confirmed cases \textit{blue} and Daily Numbers of Death \textit{green}}
    \label{fig:ground-truth-data}
\end{figure}

The selected data spreads from the 6th of March 2020 until the 24th of June 2020, covering 115 days, including the COVID-19 spring peak of the epidemic in the London area. The training dataset contains 80 days, the test set has 23 days, and the validation sample contains 12 days within the given period.

To get an insight into spatial dynamics, we chose the \textit{Facebook Population (Tile Level)}. This data stream contains information about population density at 8 am, 4 pm and midnight on a daily basis. With a view to setting a baseline model, we thought that this choice is the best compromise between a good level of spatial information about mobility, the highest level of granularity and computational complexity. The dataset for a spatial stream of information highlights the evolution of geospatial data captured periodically :
 
 \begin{itemize}
    \setlength\itemsep{0em}
    \item   Population at London Region-based level: aggregate number of people seen at a location  at 8 am, 4 pm and midnight on a daily basis and daily averaged into a framework.
\end{itemize}
 
\begin{figure}[h!tb]
    \centering
    \includegraphics[width=1\linewidth]{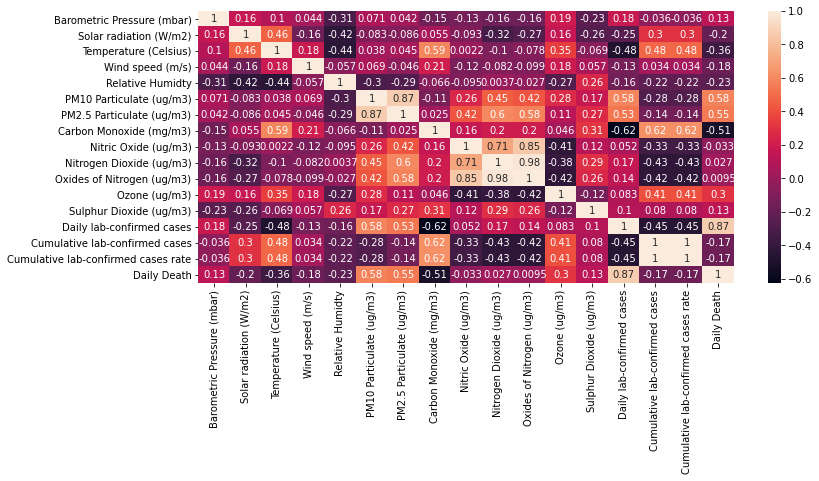}
    \caption{Correlation matrix of the temporal features}
    \label{fig:corelation-matrix}
\end{figure}

 To assess the best features for our experiments, we computed the correlation matrix of the feature and the ground truth data. Large values demonstrate significant predictive power between the variables included in this matrix. However, the absence of serious correlations may not mean necessarily mean a lack of predictive power, since only individual variables are considered for this experiment. Figure \ref{fig:corelation-matrix} shows the correlation matrix for all the temporal features analysed. 

\subsubsection{Feature Selection}
As expected, the correlation between the daily number of cases and the daily number of death is high: 0.87. Additionally, the level of PM10 and PM2.5 is significantly correlated with our ground truth data (0.58, 0.53) for the Daily lab-confirmed cases and (0.58, 0.55) for the daily death. According to the London Air Pollution Centre, these particles emanate from road traffic including carbon dioxide from engines, small metal parts and rubber from vehicle wear and braking, and road surface dirt. Others contain industrial and construction materials as well as wind-bleached dust, sea salt, pollen and soil particles. It is an indicator of intense human activity in a region. The remarkable correlation with the dynamics of COVID-19 might have a good explanation: a region where human activity is considerable means a high density of individuals, which increased the density of polluting vehicles in the area. This also leads to an increase in the potential transmission rates and risk of infection. 

\subsubsection{Feature Scaling}
Scaling features is essential before training a neural network, therefore we perform a mean normalisation. The training data is used to calibrate the mean and standard deviation for the normalisation, so scaling is performed independently of the validation and test sets. 

\subsection{Epidemiological states forecasting using Temporal stream of information}\label{sec:epidemiological-states-forecast}
CNN architectures can reproduce underlying processes and structures of the data for various image recognition tasks. Inspired by this, we explore a deep-learning framework for time series forecasting.

In time series analysis, the well-established ARMA and ARIMA models for forecasting univariate and multivariate time series are widely used. The MA part stands for Moving-Average. It indicates that the output variable depends linearly on a stochastic present and past values term. We adapted this process in order to introduce a parameter corresponding to the $WINDOW\_SIZE$ to influence the behaviour of the temporal CNN. The relative weight of the information contained in past data combined with the present information may be of interest for understanding the dynamics of the spread of a pathogen. This parameter is evaluated in our experiment in Section \ref{sec:experiements}.

\begin{figure}[htb]
    \centering
    \includegraphics[width=0.8\linewidth]{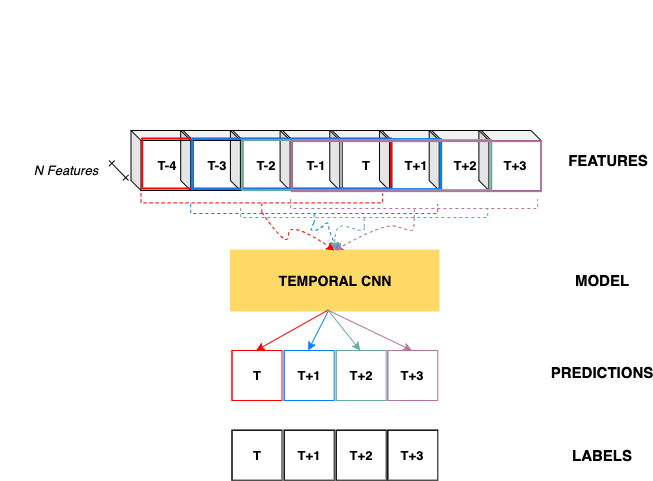}
    \caption{Time series sequence window of size 5 example}
    \label{fig:time-series-window}
\end{figure} 

This feature is implemented in our data loader function: each data loader batch request will get a sequence length window. For example, as shown in Figure \ref{fig:time-series-window}, if the batch size is set to 16, and the sequence length is 5, then the data input end up with 16 windows of length 5, each one advancing by a day. However, the architecture of the Temporal CNN depends on the window size of the sequence.

\begin{figure}[htb]
   \centering
    \includegraphics[width=1\linewidth]{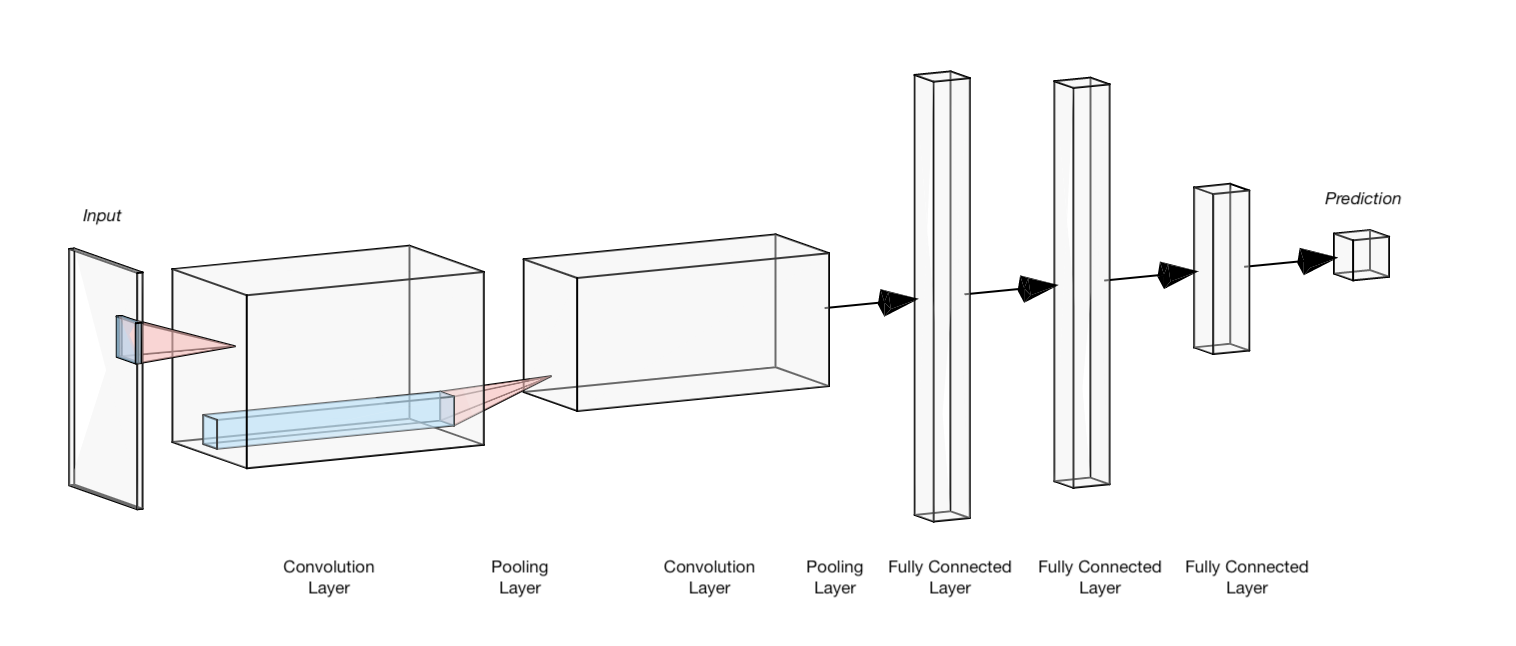}
    \caption{Temporal and Spatial CNN architecture for multivariate prediction}
    \label{fig:temporal-spatial-cnn}
\end{figure}

The baseline model is a simple CNN with two convolutional layers with pooling layers and three fully connected layers on top, illustrated in Figure \ref{fig:temporal-spatial-cnn}. The network outputs the prediction of the selected epidemiological state. The network is trained to predict the daily number of infected people and the daily number of deaths.

The input data has the shape $N \times m$, where $N$ is the length of the sequence corresponding to chosen window explained above and $m$ is the number of features. 
Convolution operations are executed on the input layer with convolution filters. A non-linear ReLu function is used for activation on each convolutional layer. The hyperparameters considered for tuning include filter size $l$ and the kernel size $K_1 \times K_2$. Given the implementation of the time-series sequence windows, a dependency between the $WINDOW\_SIZE$ and the height of the kernel $K_1$ as the kernel height cannot be greater than the window size, hence $K_1 \leq WINDOW\_SIZE$. The optimal values of parameters are determined by a grid search.

A feature map is divided into multiple segments of equal length, and each segment is then represented by its average or maximum value. The benefit of the pooling operation is that the convolutional layers output bands are down-sampled, thereby minimising uncertainty in the hidden activations. We decided to fix the filter size to 2x2, in order to reduce the dimension of the search space for the hyperparameter optimisation and avoid computation problems with the variable architecture of the CNN.

The initial time sequence, after consecutive convolutional layers and pooling layers, is characterised by a series of feature maps. The fully connected layer connects the most relevant feature maps and aggregated information from the previous convolution layers. They perform and output the regression of the desired state. The last fully connected layer has a fixed size similar to the last fully connected layer of the Spatial CNN. This is done in order to perform an equal convolution operation for the fused CNN.
 
\subsubsection{Temporal CNN architecture: Set up}

In this section, we present the results of our research into the best Temporal CNN architectures mentioned in Section \ref{sec:epidemiological-states-forecast}. Every MAE score was performed using 10 cross-validations on unseen data. We use a grid search to perform the optimal search from Table \ref{table:hyperparameters}. It provides a sample of the result with the best MAE score for the set of parameters.
\begin{table}[htb]
\centering
\begin{tabular}{llllr}  
\toprule
\multicolumn{4}{c}{Parameters} \\
\cmidrule(r){1-4}
Kernel Size 1  & Kernel Size 2  & Number of Filter 1 & Number of Filter 2 & MAE  \\
\midrule
7     & 5   & 8  & 32 &0.321 \\
5     & 5   & 8  & 32 &0.235 \\
 7        &  5       &  96 & 128 & 0.315 \\
  5        &  5       &  96 & 128 & 0.251 \\
7   &7    & 16  &  96 &0.358 \\
5 &   5 &   16  &  96    & \textbf{0.229}\\
 5    & 3   & 16  &  96    &0.269 \\

\bottomrule
\end{tabular}

\caption{Summary of the convolutional layers hyperparameters search}
\label{table:hyperparameters}
\end{table}
The value 5 for both kernels is the most appropriate. When comparing to the analysis of the convolutional filter size provided by \cite{cnn_time}, the same kernel size value results in the best prediction performance. The kernel represents the local features of the time series input or feature maps. If the size is too small, the characteristics of waveforms can not be represented. However, inversely, it will be difficult to reflect local features.

Additionally, the optimal values for the number of filters are 16 and 96. Similarly to the kernel size, the filters reflect the local features of the time series according to the above analysis. The optimal number of filters is a trade-off between enough filters to collect sufficient information from the underlined structure of the extracted data and not too much to avoid ineffective filtering.

\begin{table}[htb]
\centering
\begin{tabular}{lllr}  
\toprule
\multicolumn{4}{c}{Parameters} \\
\cmidrule(r){1-3}
Window Size & Kernel Size 1  & Kernel Size 2  & MAE  \\
\midrule
5     & 5   & 5   &0.207 \\
7     & 5   &5  &\textbf{0.104} \\
9        &  5       &  5  & 0.113 \\
12        &  5       &  5  & 0.251 \\
14   & 5    & 5   &0.182 \\
\bottomrule
\end{tabular}
\caption{Summary of the influence of the window size of the input}
\label{table:window-size}
\end{table}

We also investigate the influence of the past data combined with the present data in the forecast, by introducing a window size parameter into the Dataloader function. The results show (Table \ref{table:window-size}) that the optimal value is 7, which means that the most relevant combination of information is by taking 7 days of record before the actual day of prediction.

\subsection{Epidemiological state forecasting using a spatial stream of information}

The resolution of a spatial-temporal model for predicting the spread of an epidemic is highly correlated with the effectiveness of the local protective measure, with impact on the local economy. One effective option to model individual mobility patterns is contact tracing \cite{lorch2021quantifying}. In our case, we prefer to use aggregated data as contact tracing patterns are fairly difficult to collect and may pose privacy concerns \cite{de2020privacy}.

As pointed out in the discussion of the final feature selection in Section \ref{sec:datasets}, the daily density estimate from the \textit{Population Map (Tile Levels)} from Facebook provides a distribution of human populations for the whole London region has been selected. These maps provide statistics on the aggregate number of people seen at a location (tiles or administrative polygons) in 8-hour increments during a crisis compared to a pre-crisis baseline period. 

We acknowledge that our approach depends on the information made publicly available. However, one of our major goals is to build the most flexible model and therefore components can be swapped in or out depending on data availability. Some geographic regions might be affected by the low granularity of available data and consequently, the prediction accuracy may rely on a lower spatial-temporal resolution.

Our spatial model classifier is very similar to the temporal one: a simple CNN with two convolutional layers with pooling and four fully connected layers on top. The architecture is illustrated in Figure \ref{fig:temporal-spatial-cnn}. The network outputs the prediction of the selected epidemiological state similar to the temporal CNN.

The input data has the shape $H \times W$, where $H$ is the height of the maps and $W$ is the width of the maps. The raw maps are an array of latitude and longitude coordinates mapping the population density. We decided to map this array into a rectangular array or tensor to simulate a spatial picture of the density in London at a given time. It generates (172 $\times$ 287) tensors. The input data is pre-processed similar to the temporal data.

The number of filters is $L$ and the kernel size is $K_1 \times K_2$. The optimal values of parameters are determined by a grid search.
The non-linear ReLu function is activating each convolutional layer.

We decided to fix the filter size to 2x2, to reduce the dimension of the search space for the hyperparameter optimisation and avoid computation problems with the variable architecture of the CNN.

The four fully connected layers connect the most relevant feature maps and aggregated spatial information from the previous convolution layers. The last fully connected layer has a fixed size similar to the last fully connected layer of the Temporal CNN. This is done to perform an equal convolution operation for the fused CNN.

Epidemic dynamics are successfully emulated by various discrete states representing the health situation from different angles at a given time. For instance, several compartments, such as Susceptible, Infected, and Recovered, categorised population groups based on their sex or age to reproduce and anticipate the propagation of the infection. Ultimately, the finest classification is adopted, and the most precise the simulation is. Inspired by this approach, we decided to investigate what type of data describing the state of the epidemic is publicly available and when they are publicly released on the Internet. We only focus on publicly available data for the sake of flexibility and reproducibility of our work.

\subsubsection{Spatial CNN architecture: Set up}

For the Spatial CNN, we investigate the architecture parameters to find the optimal combination. We selected various kernel sizes and the number of filters in the grid search.

\begin{table}[htb]
\centering
\begin{tabular}{llllr}  
\toprule
\multicolumn{4}{c}{Parameters} \\
\cmidrule(r){1-4}
Kernel Size 1  & Kernel Size 2  & Number of Filter 1 & Number of Filter 2 & MAE  \\
\midrule
7    & 1   & 32  & 96 &0.152 \\
7    & 1  &  96 & 128 &0.145 \\
  11       &  1       &  96 & 128 & 0.187 \\
  11   & 3 & 32  &  96    &0.154 \\
 7   & 3 & 32  &  96    &0.147 \\
5   & 5   & 64  &  96 &0.196 \\
7 &   7 &   32  & 96   & \textbf{0.115}\\
7 &   7 &   64  & 128    & 0.128\\
 11   & 11  & 32  &  96    &0.196 \\

\bottomrule
\end{tabular}
\caption{Summary of the convolutional layers hyperparameters search}
\label{table:summary-cnn}
\end{table}

From the sample of the experiment results in Table \ref{table:summary-cnn}, $1 \times 1$ convolutions are producing interesting results. They offer a good trade-off between reducing computational load with dimensional reduction and accuracy. This operation also introduces additional non-linearity into the network \cite{lin2013network}. However, the $7 \times 7$ convolution is more robust and more stable on this data according to the experiment.

For the selected input features (172 $\times$ 287), it is not always relevant to increase the number of convolutional filters, as they will increase the redundancy without a significant impact on the final prediction while increasing the computation cost. The optimal number was respectively 32 and 96. 

\subsection{Combined CNN Forecast}
 In this section, we are comparing the Temporal and Spatial baseline model with the Fused-CNN based on the combination exposed in section \ref{sec:fusedcnn} We adopted two commonly used CNN architectures for temporal and spatial network, pre-trained with their optimal parameters identified previously.  The Fused-CNN has been trained with its optimal parameters found in the same manner as the CNN-T and CNN-S. All the assessments of MAE scores have been done using 10 cross-validations.

\begin{table}[!htb]
\centering
\begin{tabular}{llr}  
\toprule
Model & MAE & Relative Improvement of the Fused-CNN  \\
\midrule
$CNN-T_{both}$& 0.107& 28.4 \% \\
$CNN-S_{both}$   & 0.122   & 37.2 \% \\
Fused-CNN    & \textbf{0.0766}  &  \textbf{Avg. 32.8 \%} \\
\bottomrule
\end{tabular}
\caption{Summary of the Fused-CNN evaluation}
\label{table:summary-fused-CNN}
\end{table}

Table \ref{table:summary-fused-CNN} shows the performance for the three networks implemented, demonstrating improvements for an average of 32.8\% accuracy. The fusion operation is potentially avoiding overfitting given that only the pairs of features that the two networks agreed on can contribute to the regression. All the features with the potential for overfitting were dropped off.

\subsection{Performance evaluation of combining multiple data sources}
In this section, we explore the Data Assimilated step implemented to forecast the state of the model, based on the description provided in Section \ref{sec:data-assimilation-theory}.

Sequential Bayesian filters are sensitive to initialisation ensemble. To infer the sensitivity of our filter, multiple initialisations of the state model have been explored. It is fairly known that the stochastic Kalman Filter introduces stochastic noise. This could affect the performance of the filter to the extent that it depends on the exact configuration of the DA step, on the nonlinearity of the dynamics, and, above all, on the size of the ensemble \cite{book}. 

The increase in the variance of the ensemble perturbation improved the convergence of the filter. Table \ref{table:summary-enkf} summarised the key information.

\begin{table}[htb]
\centering
\begin{tabular}{ccccc}  
\toprule
\multicolumn{3}{c}{Parameters} \\
\cmidrule(r){1-3}
State initialisation & R: Observation Covariance Matrix & Ensemble Size & MAE  \\
\midrule
Zero   & 0.01   & 50  & 0.082 \\
Previous State   & 0.01  &  50 & 0.114 \\
  One       &  0.01       &  50 & 0.212  \\
Zero   & 0.1   & 50  & \textbf{0.074}   \\
Previous State   & 0.1  &  50 & 0.094 \\
  One       &  0.1       &  50 & 0.081  \\
Zero   & 0.1   & 100  & 0.121 \\
Previous State   & 0.1  &  100 & 0.132 \\
  One       &  0.1       &  100 & 0.086 \\

\bottomrule
\end{tabular}
\caption{Summary of the EnKF forecast parameters search}
\label{table:summary-enkf}
\end{table}

The best set of parameters is 0.1 for the Observation Covariance Matrix $R$, an ensemble size of 50, a value commonly seen in the literature, and zero-initialised ensemble states. The variance of the ensemble perturbation was fixed at 1.

\subsection{Architecture Flexibility}

In this last experiment, we are testing the flexibility of our approach by removing initial states and models to check the relative improvements to the default model. For example, we tested the prediction of a single epidemiological state: $CNN-T_{cases}$ is predicting the daily number of COVID-19 cases, $CNN-T_{deaths}$ - the daily number of deaths and $CNN-T_{cases,deaths}$ is predicting both.

Model learning capabilities are depending on the state label and the features. As seen in Table \ref{table:summary-cnn-s-evaluation}, the $CNN-T_{deaths}$ is more likely to extract and select relevant patterns from the features than the $CNN-T_{cases}$ does. Additionally, it can be inferred that some intrinsic characteristics of both states are shared across layers as the simultaneous prediction of the two states in an average of the single state predictions. The more various states are predicted, the more information is cross-correlated and shared across the layers and potentially, the more robust and accurate the prediction might be. Especially when some relevant states labels are missing due to a lack of available data, the other states might be interested in activating other internal connections between features to emulate the dynamics of the infection correctly.

\begin{table}[!htb]
\centering
\begin{tabular}{cc}  
\toprule
Model & MAE   \\
\midrule
$CNN-T_{cases}$ &0.112 \\
$CNN-T_{deaths}$       &0.091\\
$CNN-T_{cases,deaths}$ &0.107 \\
\midrule
$CNN-S_{cases} $   & 0.148 \\
$CNN-S_{death} $      &0.235 \\
$CNN-S_{cases,deaths}$ &\textbf{0.122}\\

\bottomrule
\end{tabular}
\caption{Summary of the evaluation of CNN-T and CNN-S multiple state prediction flexibility.}
\label{table:summary-cnn-s-evaluation}
\end{table}

 Contrary to the Temporal CNN, forecasting the number of cases is more precise than forecasting death. The $CNN-S_{cases}$ seems to learn better the underlying infection dynamics Table \ref{table:summary-cnn-s-evaluation}. We understand as the input data are a daily frame of the population density in London, the difference in population density between two successive frames is somewhat related to population movement. This movement could perpetuate the virus transmission through human contact. With multiple time frame analyses performed by the convolutional layers, the algorithm could infer this notion of transmission and, consequently, better predict the number of cases as it is directly linked with it. Additionally, it could be possible that the algorithm analyses the hottest point on the map to directly map an increase in the case directly counts, considering these points as areas of potential virus transmission. Granted, it does not take into consideration social distancing and hygiene measures. However, this information might be included in future work.

\section{Results and Discussion}

\subsection{Evaluation of noise insensitivity}

The combination of CNN shows its ability to combine knowledge from the Temporal and Spatial CNN. Even if the previous experiments highlight an attractive characteristic: the CNN-T is better at predicting the number of states and the CNN-S the number of death, both networks agreed the feature maps at the exact location (with probably different intensity), if not, the performance would not have been increased. It is probable that without the other network, the activation is not intense enough to be activated. Hence, the knowledge is well shared, and identical parts of epidemic dynamics are indirectly emulated.

\begin{figure}[htb]
    \centering
    \includegraphics[width=0.5\linewidth]{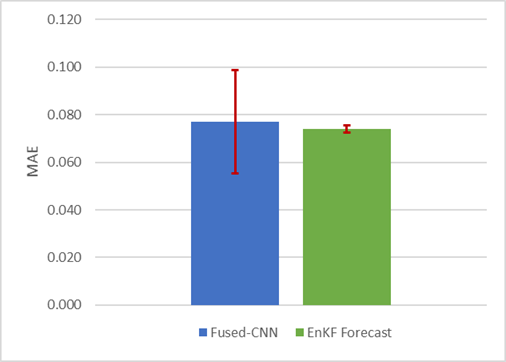}

\caption{Summary of the final EnKF forecast evaluation }
\label{figure:enkf_forecast}
\end{figure}

The EnKF forecast provides minimal improvement (Figure \ref{figure:enkf_forecast}). However, from the previous experiments, incorporating uncertainties into the state provides robustness and consistency during the forecast step proceeded by the Fused-CNN. This is particularly interesting when forecasting states from multiple streams of information.  

\subsection{Comparison with compartmental models}

We compared our model to a variety of compartmental models. The first model, SEIR \cite{li1995global}, is a standard compartmental model in which the population is divided into Susceptible (S), Exposed (E), Infectious (I), and Recovered (R) individuals. 
The second comparison model is an extended SEIR model \cite{reno2020forecasting}, introducing checkpoints to change some model parameters during the simulation.
The checkpoints introduced in this model are closely linked to the evolution of government action regarding the surge of Covid-19 in the UK. They correspond to starting a form of social distancing with the general lockdown announcement (On March 23rd 2020) and then stopping social distancing (On May 10th 2020).
These two standard compartment models capture significant aspects of infectious disease dynamics. Still, they are deterministic mean-field models that assume uniform population mixing (every individual in the population is equally likely to interact with every other individual). 
Then, we decided to compare our model against a network SEIR model \cite{SMIRNOVA2019}. When investigating disease transmission, it is often essential to investigate disease transmission using stochasticity, heterogeneity, and the structure of contact networks, where trying to limit the spread can be viewed as perturbing the contact network (e.g. social distancing) or making use of it (e.g. contact tracing). 

\begin{figure}[htb]
    \centering
    \includegraphics[width=0.5\linewidth]{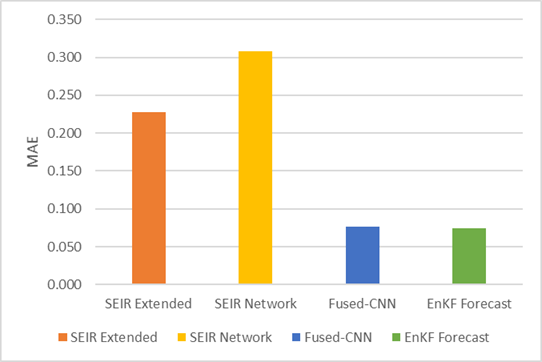}

\caption{Comparison of our proposed EnKF model with various variants of the SEIR models }
\label{figure:seir}
\end{figure}

All the parameters are inferred from the recent COVID-19 pandemic outbreak in Wuhan \cite{LIN2020211}. The population is the size of the Greater London region at the starting day of our experiment, on March 2nd 2020). Figure \ref{figure:seir} shows the results of the comparative experiments.

Compared to the SEIR models, additional data sources can be beneficial in reducing uncertainty. However, they can also add extra noise, which can bias results significantly. In our model (EnKF), using Data Assimilation, a combination of multiple data sources and a fused CNN approach overcomes the issue.

The Extended SEIR model is designed to track behavioural changes in the population, helping it make better overall predictions. Our model automatically learns these patterns from the data through the social media and air quality dataset, which can be used as proxies for human activity indicators. Regarding COVID-19, these indicators tend to capture when people started social distancing, strongly related to quarantine and self-isolation measures governments took. 

EnKF also offers superior flexibility compared to traditional SEIR models, as it is purely data-driven and built upon a modular architecture. When a data source stops being useful to make predictions or is used as a proxy for human activity, the stream can be replaced. A new CNN architecture can be trained, and the output would be fused with the other CNN models available. 

\section{Conclusion}

This paper introduces a novel approach to predicting epidemiological parameters by integrating daily signals from various sources of information. The combination of CNN models from different data streams has successfully been implemented to build robust predictions to simulate the parameters of the infection dynamic. The proposed approach outperforms standard CNN predictions by an increase of up to 32.8\% accuracy on COVID-19 temporal and spatial data. The Data Assimilation step also introduces stronger robustness and consistency to our prediction.

There are a few avenues to explore in this research project. One of the most noticeable additions is integrating other data streams with different modalities, sample rates and timelines. For instance, additional social media information, such as aggregated message counts for various, can be a good proxy for human activity, especially in strict lockdown periods. In the London Greater Area, Transport for London offers near real-time activity monitoring on all transport links: tube, buses, trains and cycling information, which is another good predictor of how well the lockdown is enforced. If given access to a stream of private datasets, such as card payments or supermarket aggregated delivery information, this model could show that the ensembles' flexibility is a powerful predictor.

Moreover, our approach may require further assessment and validation on larger datasets. This could be done by assessing the flexibility and adaptability of a new city or pathogen.

\section*{Author contributions}

\noindent\textbf{Romain Molinas:} Conceptualization, Methodology, Software, Formal analysis, Data Curation, Writing - Original Draft, Visualization; \textbf{César Quilodrán Casas:} Conceptualization, Methodology, Writing - Review \& Editing, Supervision; \textbf{Rossella Arcucci:} Methodology, Writing - Review \& Editing; \textbf{Ovidiu \cb Serban:} Conceptualization, Methodology, Writing - Original Draft, Supervision.

\section*{Competing interests}
\noindent The authors have no competing interests to declare.

\bibliographystyle{elsarticle-num}
\bibliography{references}
\end{document}